\documentclass[11pt]{article}
\usepackage{graphicx}
\usepackage{algorithm}
\usepackage{algorithmic}

\usepackage{caption}
\usepackage{subcaption}

\usepackage{ACL2023}

\usepackage{times}
\usepackage{latexsym}

\usepackage[T1]{fontenc}

\usepackage[utf8]{inputenc}

\usepackage{microtype}

\usepackage{inconsolata}

\title{EvolveMT: an Ensemble MT Engine Improving Itself with Usage Only}

 \author{Kamer Ali Yuksel, Ahmet Gunduz, Mohamed Al-Badrashiny,\\ {\bf Shreyas Sharma}, {\bf and Hassan Sawaf} \\
         aiXplain Inc., 16535 Grant Bishop Lane, Los Gatos, CA 95032, US \\ \{kamer, ahmet, mohamed, shreyas,  hassan\}@aixplain.com}

\begin{document}
\maketitle
\begin{abstract}
This paper presents EvolveMT for efficiently combining multiple machine translation (MT) engines. The proposed system selects the output from a single engine for each segment by utilizing online learning techniques to predict the most suitable system for every translation request. A neural quality estimation metric supervises the method without requiring reference translations. The online learning capability of this system allows for dynamic adaptation to alterations in the domain or machine translation engines, thereby obviating the necessity for additional training. EvolveMT selects a subset of translation engines to be called based on the source sentence features. The degree of exploration is configurable according to the desired quality-cost trade-off. Results from custom datasets demonstrate that EvolveMT achieves similar translation accuracy at a lower cost than selecting the best translation of each segment from all translations using an MT quality estimator. To our knowledge, EvolveMT is the first meta MT system that adapts itself after deployment to incoming translation requests from the production environment without needing costly retraining on human feedback. 
\end{abstract}

\section{Introduction}
Machine Translation (MT) has experienced substantial progress in recent years, resulting in improving accuracy and more human-like translation output. Despite these advancements, challenges remain, particularly in ensemble modeling. Ensemble models integrate predictions from multiple individual models to achieve a more accurate final output. However, the effective combination of these models is often a complex task that requires thoughtful consideration of factors such as the model architecture, training data, and prediction combination methods. One of the significant challenges in MT ensembling is that the training data used to train the ensemble model, may not be fully representative of the data to be translated later, leading to a mismatch between the model and the data. This paper presents EvolveMT, a method that addresses data drift in ensemble models by continual self-adaptation for optimal performance during usage.

In the subsequent section, we review existing machine translation (MT) quality estimation metrics in the literature, which have been trained on human evaluation or post-editing datasets. In the Approach section, we present a comprehensive explanation of the proposed method. In the Experiments section, we describe our experimental design and provide quantifiable results demonstrating the enhancement resulting from the application of the proposed method, as compared to state-of-the-art quality estimation metrics. Finally, we discuss the obtained results and present our conclusions.

\section{Related Work}
In the WMT20 Metrics Shared Task \citep{mathur2020results}, four reference-free metrics were submitted to evaluate machine translation outputs in the news translation task. These metrics use bilingual mapping of contextual embeddings from language models such as XLM-RoBERTa \citep{conneau2019unsupervised} to assess cross-lingual semantic similarity. However, they often struggle to accurately differentiate between human and machine translations, except for COMET-QE \citep{rei-etal-2020-comet}, the only reference-free metric capable of doing so.

The study by \citet{freitag2021experts} evaluated top MT systems from WMT 2020 using Multidimensional Quality Metrics (MQM) and professional translator annotations. Their results showed a low correlation between crowd worker evaluations and MQM, leading to different rankings and questioning previous conclusions. The study also found that automatic metrics based on pre-trained embeddings can outperform human crowd workers, suggesting that models trained with crowd-sourced human evaluations may have higher accuracy.

The WMT21 Metrics Shared Task \citep{freitag2021results}, used MQM expert-based human evaluation to acquire reliable ratings, and evaluate metrics on news and TED talk translations produced by MT systems. Results showed reference-free metrics COMET-QE and OpenKiwi performed well in scoring human translations but not as well with MT outputs, and were strong at segment-level human translation evaluation while competitive with reference-based metrics in system-level evaluation. 

REGEMT \citep{vstefanik2021regressive} is a reference-free metric in WMT21 that uses an ensemble of surface, syntactic, and semantic similarity metrics as input to a regression model. As demonstrated by CushLEPOR \citep{han2021cushlepor}, it allows customization, outperforming lexical semantic similarity-based metrics with a higher computational cost. 

Onception \citep{mendoncca2022onception} used active learning to converge an MT ensemble in a production environment to the best MT with evaluations acquired online. 

\citep{naradowsky2020machine} used bandit-learning to adapt MT policies based on simulated user feedback, outperforming the best single MT in mixed-domain settings. A contextual bandit strategy was proposed to make instance-specific decisions, but the system still required a human-in-the-loop (HITL) process.

\section{Approach}
EvolveMT is a quasi MT ensemble technique. In contrast to the traditional multi-system MT approach, which combines outputs from multiple MT systems to enhance translation accuracy and fluency, EvolveMT prioritizes the selection of the most optimal translation from a finite set of MT systems, as we demonstrate in this section. Figure \ref{fig:diagram} below shows the system architecture of EvolveMT.  The system is centered around a multi-class classification model that drives multiple processes to select the best MT model for translation requests. 

For each incoming machine translation request,  we use SpaCy \citep{spacy2} and Stanza \citep{qi-etal-2020-stanza} frameworks to extract morphological and lexical features. These features include the count of tokens, characters, and the average word length, as well as the frequency of Part-of-Speech labels (such as nouns, verbs, adjectives, etc.), the frequency of Named Entity Recognition labels (including entities such as persons, locations, organizations, etc.), and the frequency of morphological features (e.g. gender and aspect). These features are combined with the 1024-dimensional embedding vector generated by the XLM-RoBERTa encoder of the COMET-QE model and stored alongside the input sentence in the \emph{Ranked Batch Requests Queue}. This queue serves the purpose of prioritizing translation requests that necessitate precedence in processing.

At the outset, requests in the Ranked Batch Requests Queue are ranked based on the order in which they are added. The highest-ranked Machine Translation (MT) request is selected for translation. The \emph{Multi-class MT Classifier} employs the extracted features of the selected MT request to determine the MT systems to be utilized. The classifier prioritizes MT systems with a higher probability of having a higher COMET-QE value. Exploration of additional MT systems becomes more likely only if the probabilities from the classifier's prediction exhibit high entropy. This enables EvolveMT to minimize the cost of exploration when the best MT is predicted with high certainty. 

Finally, the selected MT systems are utilized to translate the MT request, and the COMET-QE score is calculated for each translation. The translation with the highest score is chosen and returned in response to the MT request. The Multi-class MT Classifier is then updated online with the best MT system, as determined by the COMET-QE score, serving as a label for the extracted features of the MT request. The online machine learning (ML) functionality of FLAML AutoML framework \citep{wang2021flaml} is utilized for online learning. This capability enables the optimization of model hyper-parameters during the iterative course of ML, facilitating continual ML without repetitively hyper-tuning the classifier from scratch.

Subsequently, the classifier is employed to re-rank the Ranked Batch Requests Queue, based on the uncertainty of the classifier, with requests having higher entropy being placed at the top of the queue for prioritized translation. Getting the MT request with the maximum entropy from the queue after each iteration, helps prioritize the most informative sample for the iterative training of the classifier. As the classifier improves its ability to predict the best MT model for MT requests via learning, it reduces the likelihood of exploring other MT(s).

The driving algorithm in EvolveMT, which outlines the primary process of the proposed method, is presented in pseudo-code form in Algorithm \ref{alg:cap}.

\begin{algorithm}[h]
\caption{EvolveMT with online active-learning}
\label{alg:cap}
\begin{algorithmic}
\algsetup{linenosize=\tiny}
\scriptsize
\REQUIRE $MTQueue$: list of tuples (source text, features), \newline and $MaxMTs$: maximum number of MTs to sample
\WHILE{$len(MTQueue)$}
\STATE $Classifier.rankByUncertainty(MTQueue)$
\STATE $source, feats \gets MTQueue.popMaxEntropyItem()$
\STATE $predMT, classProbs \gets Classifier.predict(feats)$
\STATE $predTrans \gets Translate(source, predMT)$
\STATE $randMTs \gets sampleMTs(classProbs, MaxMTs)$
\STATE $maxEnt \gets normalizedEntropy(classProbs)$
\IF{$randMTs_{0} = predMT$ \AND $maxEnt < \alpha$} 
\STATE $Classifier.learn(feats, predMT)$
\ELSE
\STATE $sampled \gets Translate(source, randMTs)$
\STATE $randScores \gets CometQE(source, sampled)$
\STATE $predMTScore \gets CometQE(source, predTrans)$
\IF{$max(randScores) >predMTScore$}
\STATE $Classifier.learn(feats, randMTs)$
\STATE $IndexOfBestMT \gets randScores.argMax()$
\STATE $predTrans \gets sampled_{IndexOfBestMT}$
\ELSE
\STATE $Classifier.learn(feats, predMT)$
\ENDIF
\ENDIF
\STATE $respondMTRequest((source, predTrans))$
\ENDWHILE
\end{algorithmic}
\end{algorithm}

\begin{figure}[H]
\includegraphics[width=\columnwidth]{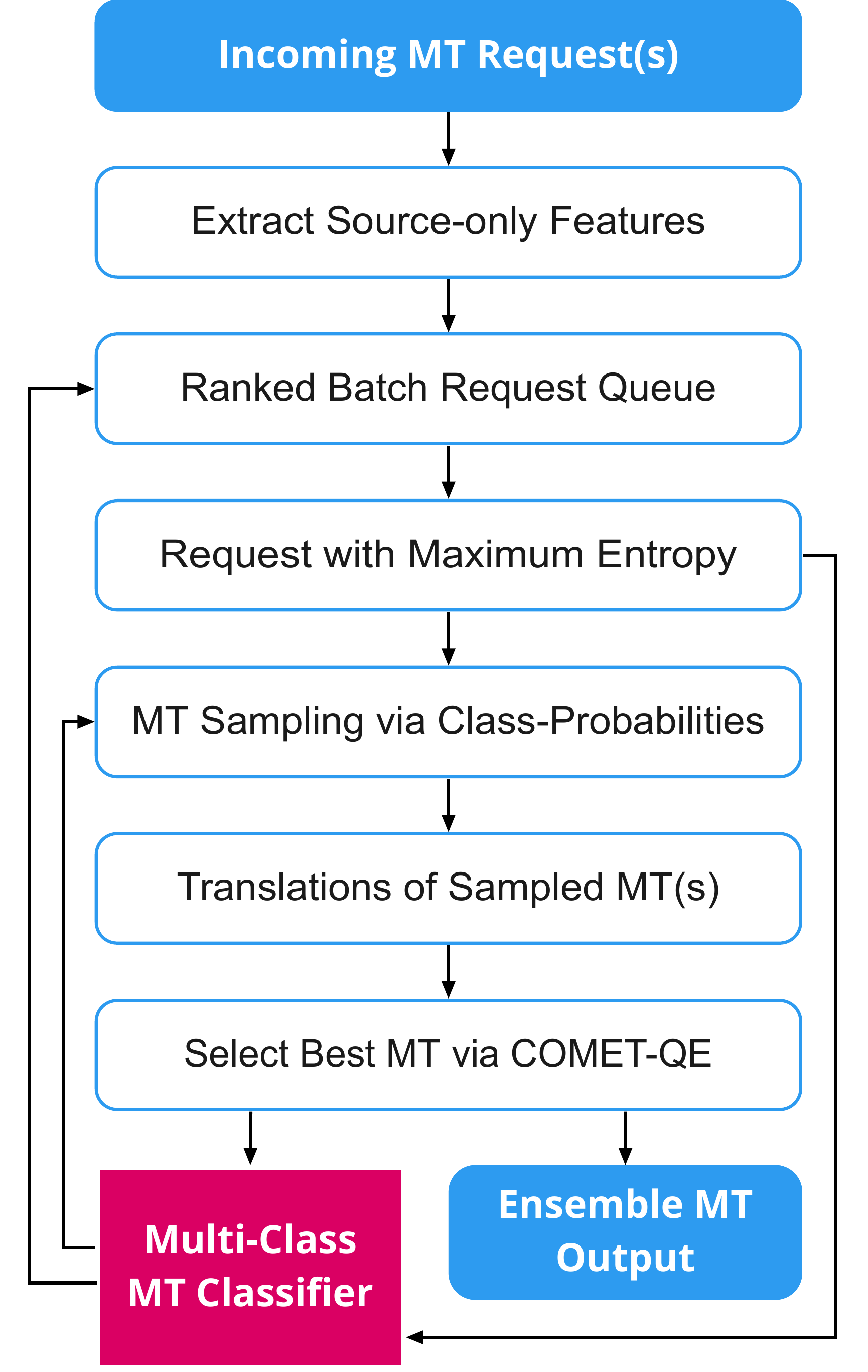}
\caption{EvolveMT System Architecture}
\label{fig:diagram}
\end{figure}

\newpage

\section{Experiments}
\subsection{Data}
A multi-lingual corpus of 37,500 human-translated sentences in Czech, German, and Russian, along with their corresponding English source-texts, was collected for the OPUS repository \citep{lison-tiedemann-2016-opensubtitles2016, aulamo-tiedemann-2019-opus} using stratified random sampling for each language and dataset. To evaluate EvolveMT, translations for each sentence in the corpus were obtained from one open-source machine translation system \citep{https://doi.org/10.48550/arxiv.2212.01936} and five major machine translation service providers in the industry (Google, Azure, AWS, ModernMT, and AppTek).

\subsection{Setup}
Experiments were conducted on a 64-bit Ubuntu 22.04 LTS computer system with an AMD Ryzten 5950X CPU (16 processors, 32 threads) and 64GB of memory. An Nvidia GeForce RTX GPU was used for XLM-RoBERTa embedding extraction from the fine-tuned COMET-QE encoder. The results showed an average 2.88 ($\pm$0.06) millisecond response time for the EvolveMT system to return an MT output and update its classifier when the MT request queue contained a single item. Depending on GPU usage, the feature extraction time was (183.43 -244.80) milliseconds. It's worth noting that in a production setting, feature extraction can be performed in parallel for multiple MT requests.

\subsection{Evaluation criteria and baselines}
In this paper, grid search is used to evaluate the impact of two hyperparameters, $maxMTs$ and $\alpha$, on the classifier's performance. $maxMTs$ refers to the maximum number of machine translation systems the classifier can select, and $\alpha$ is the maximum entropy threshold (as described in Algorithm \ref{alg:cap}). The grid search involves varying $maxMTs$ from 1 to 6 (the maximum number of the MT systems we are using), and $\alpha$ from 0.1 to 1.0 with increments of 0.1. The experimental results are obtained by averaging 100 repetitions to account for the method's inherent stochasticity. For clarity in the results section, we present the results over the $maxMTs$ range while setting $\alpha$ to its optimal value of 0.2, determined through the grid search. 

For the evaluation, we adopt the reference-based quality score COMET-DA, as detailed in \citep{rei-etal-2020-comet}, as the evaluation metric for our ensemble output. This choice is motivated by the results of prior research which have demonstrated that COMET-DA exhibits a higher correlation with human evaluation scores compared to other widely used machine translation metrics, such as BLEU and METEOR. The evaluation of EvolveMT is being conducted against the following baselines:

\begin{itemize}
  \item COMET-QE Ensemble: translation is performed utilizing the six MT systems. The translation with the highest COMET-QE score is selected for each input sentence as the ensemble translation. Then, the COMET-DA score is calculated using selected translations.
  \item Best MT: involves translating the entire data using all six MT systems. The MT system that produces the highest overall COMET-DA is then selected as the Best MT to employ.
\end{itemize}

\subsection{Results}
The comparison of COMET-DA scores of the COMET-QE ensemble and Best MT concerning various variants of EvolveMT with varying $MaxMTs$ values are presented in Table \ref{tab:costs}. The results are depicted for the three language pairs of English-to-Czech, English-to-German, and English-to-Russian. In addition, the average translation cost of each system across the three languages is also documented in the table. The findings indicate that EvolveMT approximates the COMET-QE ensemble's quality while incurring significantly lower costs. 

Furthermore, the results in the table reveal that the optimal cost-quality trade-off for EvolveMT varies depending on the target language. Specifically, for all three language pairs, it can be observed that EvolveMT with $MaxMTs=3$ and $MaxMTs=4$ provide the best balance between cost and quality when compared to other individual and ensemble MTs. As $MaxMTs$ increases, EvolveMT can achieve higher MT quality by exploring a larger pool of MT systems from which the best translation can be selected. Hence, the $MaxMTs$ parameter can be adjusted to achieve the desired cost-quality trade-off.

\begin{figure}
\includegraphics[width=\columnwidth]{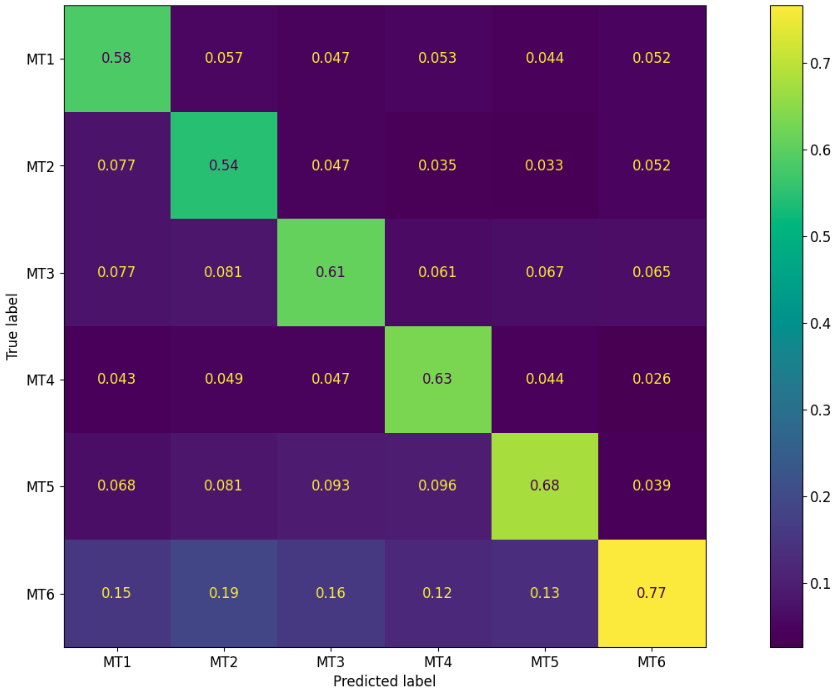}
\caption{The normalized confusion matrix between EvolveMT ($MaxMTs$ = 4) and COMET-QE after 100 translations requests.}
\label{fig:cmatrix}
\end{figure}

Notably, after only a few hundred Machine Translation (MT) requests from the total dataset, the EvolveMT algorithm demonstrates convergence towards an upper limit of its weighted F1-score, which depends on the parameter $maxMTs$. Figure \ref{fig:cmatrix} shows the confusion matrix between the outputs of EvolveMT (with $MaxMTs$ = 4) and the COMET-QE ensemble after 100 translations requests. The swift convergence of EvolveMT with a limited number of requests is mainly due to the utilization of XLM-RoBERTa embeddings that have been fine-tuned specifically for the COMET-QE task. This exemplifies the model's effectiveness, as it begins with no prior knowledge, and within a few hundred requests, it can converge and approach the performance of the COMET-QE ensemble. It is crucial to mention that the results presented in Table \ref{tab:costs} encompass the "warm up" phase where EvolveMT starts from zero knowledge until full convergence is achieved. If this phase were excluded, the COMET-DA scores of EvolveMT would likely be even higher.

\begin{table*}[ht]
\centering
\begin{tabular}{l|l|c|c|c}
\hline
             &        & \multicolumn{3}{c}{\textbf{COMET-DA}}                          \\ \hline
\multicolumn{1}{c|}{\textbf{Model}} &
  \multicolumn{1}{c|}{\textbf{Cost (\$)}} &
  \multicolumn{1}{c|}{\textbf{English-to-Czech}} &
  \multicolumn{1}{c|}{\textbf{English-to-German}} &
  \textbf{English-to-Russian} \\ \hline
Best MT (1)     & 20.000 & \multicolumn{1}{c|}{0.867} & \multicolumn{1}{c|}{0.586} & 0.617 \\
COMET-QE (6)     & 77.000 & \multicolumn{1}{c|}{0.900} & \multicolumn{1}{c|}{0.605} & 0.658 \\ \hline
EvolveMT (1) & 12.312 & \multicolumn{1}{c|}{0.851} & \multicolumn{1}{c|}{0.567} & 0.605 \\
EvolveMT (2) & 23.442 & \multicolumn{1}{c|}{0.870} & \multicolumn{1}{c|}{0.586} & 0.627 \\
EvolveMT (3) & 32.358 & \multicolumn{1}{c|}{0.878} & \multicolumn{1}{c|}{0.591} & 0.637 \\
EvolveMT (4) & 39.905 & \multicolumn{1}{c|}{0.882} & \multicolumn{1}{c|}{0.596} & 0.643 \\
EvolveMT (5) & 46.067 & \multicolumn{1}{c|}{0.887} & \multicolumn{1}{c|}{0.598} & 0.647 \\
EvolveMT (6) & 51.095 & \multicolumn{1}{c|}{0.887} & \multicolumn{1}{c|}{0.599} & 0.651 \\ \hline
\end{tabular}
\caption{The Cost and COMET-DA comparison of the Best MT system, COMET-QE and EvolveMT ensembles for various $MaxMTs$ parameters (indicated in parentheses). The MT quality increases as COMET-DA scores increase}
\label{tab:costs}
\end{table*}

\section{Discussion}

The cost-benefit analysis of EvolveMT highlights the trade-off between run-time efficiency and training expenses. While the run-time cost of EvolveMT may be higher than that of Best MT, it does not require the extensive and time-consuming training process required for traditional MT ensemble methods. This training process involves obtaining translations from all MTs and scoring them using references generated by annotators.

However, the increased run-time cost of EvolveMT is offset by its ability to achieve superior production quality and adapt to changes in the data domain with a minimum amount of overhead. As the data domain changes, traditional MT ensemble techniques require costly retraining to accommodate the new domain, whereas EvolveMT can adapt to changes with a few hundred MT requests.

This versatility and adaptability of EvolveMT make it a robust solution for machine translation tasks that may be subject to data variation, as it can adjust to these changes with minimal effort. The cost-benefit analysis results clearly demonstrate that the increased run-time cost of EvolveMT is outweighed by its high performance and adaptability in the face of changing data domains.

\section{Limitations}

The performance of the EvolveMT system is contingent upon the reliability of the COMET-QE model in providing accurate labels for the MT requests. Utilizing the encoder's embeddings as features necessitates that the COMET-QE model performs effectively on blind MT requests. The batch re-ranking of MT requests after each learning step may result in a computational bottleneck if the queue size is substantial. To mitigate this issue, an asynchronous re-ranking process could be implemented, whereby the queue is only reorganized once the re-ranking is completed. Additionally, before the re-ranking process, a diverse subset of the queue can be selected based on the XLM-RoBERTa embeddings, which reflect the novelty of the requests relative to previously processed MT requests. The source embeddings from the XLM-RoBERTa model can be cached in parallel during the batch feature extraction process utilizing GPU capabilities, thus facilitating efficient COMET-QE inference. EvolveMT could also be optimized for cost-effectiveness by incorporating the cost of each MT in the ensemble into the algorithm.

\section{Conclusion and Future Work}

This study presents a novel approach called EvolveMT for ensembling machine translation (MT) engines, focusing on minimizing the number of engines required to be queried to achieve optimal quality. To evaluate the efficacy of the proposed method, a series of experiments were conducted, wherein EvolveMT was implemented with varying levels of granularity in terms of the maximum number of engines permitted for each individual MT request. The quantitative results of the experiments indicate that, compared to the traditional method of querying all available MT engines, EvolveMT offers a more cost-effective solution for the ensembling process without compromising the quality of the resulting translations.

EvolveMT presents a unique advantage in terms of cost efficiency compared to COMET-QE Ensemble. This is achieved by utilizing a stochastic exploration approach that selectively queries additional MT engines based on predicted probabilities, which are also employed in an active-learning framework by re-ranking MT requests after each learning step. Furthermore, unlike traditional MT ensemble techniques, EvolveMT can adapt in real-time to changes in customers' translation requests, without incurring the cost of acquiring human references or undergoing costly re-training or fine-tuning.

In conclusion, this paper presents four significant contributions to the field of machine translation: (1) the introduction of the first self-improving MT system that operates without the need for human feedback; (2) the capability of adaptively optimizing the MT ensemble in response to production environment translation requests through online machine-learning; (3) the development of a novel approach for selectively querying MT engines rather than relying on translations from all available engines; and (4) the implementation of an active-learning framework that leverages uncertainties from the ensemble for batch translation. 

\section{Ethics and Impact Statement}
EvolveMT is a high-quality machine translation (MT) system for individuals or organizations. It can improve translation accuracy if validated on a specific MT corpus. EvolveMT is trained from scratch for each customer or project, eliminating biases in the algorithm but may still present biases in the quality estimation metric or training dataset. The system is self-adaptable, secure, and protects user privacy by deleting data immediately after translation. EvolveMT eliminates the need for re-training and re-hypertuning, reducing computational costs and being environmentally friendly. The only potential harm is to linguists who perform post-editing as it reduces their dependence on references or evaluations.

\bibliography{ACL2023Industry}
\bibliographystyle{acl_natbib}

\end{document}